\documentclass[letterpaper, 10 pt, conference]{ieeeconf} 
\usepackage{subcaption}
\usepackage{graphicx}
\usepackage{changepage}
\usepackage{booktabs}

\IEEEoverridecommandlockouts 

\usepackage{amsmath} 

\title{\LARGE \bf
Lander.AI: Adaptive Landing Behavior Agent for Expertise in 3D Dynamic Platform Landings
}

\author{Robinroy Peter, Lavanya Ratnabala, Demetros Aschu, Aleksey Fedoseev, and Dzmitry Tsetserukou
\thanks{The authors are with the Intelligent Space Robotics Laboratory, Skolkovo Institute of Science and Technology, Bolshoy Boulevard 30, bld. 1, 121205, Moscow, Russia}
\thanks{
email: {(robinroy.peter, lavanya.ratnabala, demetros.aschu, aleksey.fedoseev, d.tsetserukou})@skoltech.ru}}

\begin{document}

\maketitle
\thispagestyle{empty}
\pagestyle{empty}


\begin{abstract}
Mastering autonomous drone landing on dynamic platforms presents formidable challenges due to unpredictable velocities and external disturbances caused by the wind, ground effect, turbines or propellers of the docking platform. This study introduces an advanced Deep Reinforcement Learning (DRL) agent, Lander.AI, designed to navigate and land on platforms in the presence of windy conditions, thereby enhancing drone autonomy and safety. Lander.AI is rigorously trained within the gym-pybullet-drone simulation, an environment that mirrors real-world complexities, including wind turbulence, to ensure the agent's robustness and adaptability. 

The agent's capabilities were empirically validated with Crazyflie 2.1 drones across various test scenarios, encompassing both simulated environments and real-world conditions. The experimental results showcased Lander.AI's high-precision landing and its ability to adapt to moving platforms, even under wind-induced disturbances. Furthermore, the system performance was benchmarked against a baseline PID controller augmented with an Extended Kalman Filter, illustrating significant improvements in landing precision and error recovery. Lander.AI leverages bio-inspired learning to adapt to external forces like birds, enhancing drone adaptability without knowing force magnitudes.This research not only advances drone landing technologies, essential for inspection and emergency applications, but also highlights the potential of DRL in addressing intricate aerodynamic challenges.
\end{abstract}

\section{Introduction}

The field of autonomous drone navigation has experienced significant advancements in recent years, propelled by rapid progress in artificial intelligence and robotics. Unmanned Aerial Vehicles (UAVs) have become increasingly vital across a range of applications, including surveillance, delivery services, environmental monitoring, and emergency response \cite{b0}. A pivotal aspect of these applications is the drone's ability to perform precise and safe landings on moving platforms a task that continues to pose substantial challenges. This challenge is further intensified by complex wind forces \cite{b26}. 

Traditional control methods often struggle to dynamically adapt to these rapidly changing aerodynamic conditions \cite{b8}. 
\begin{figure}[ht]
 \centering
 \includegraphics[width=0.48\textwidth]{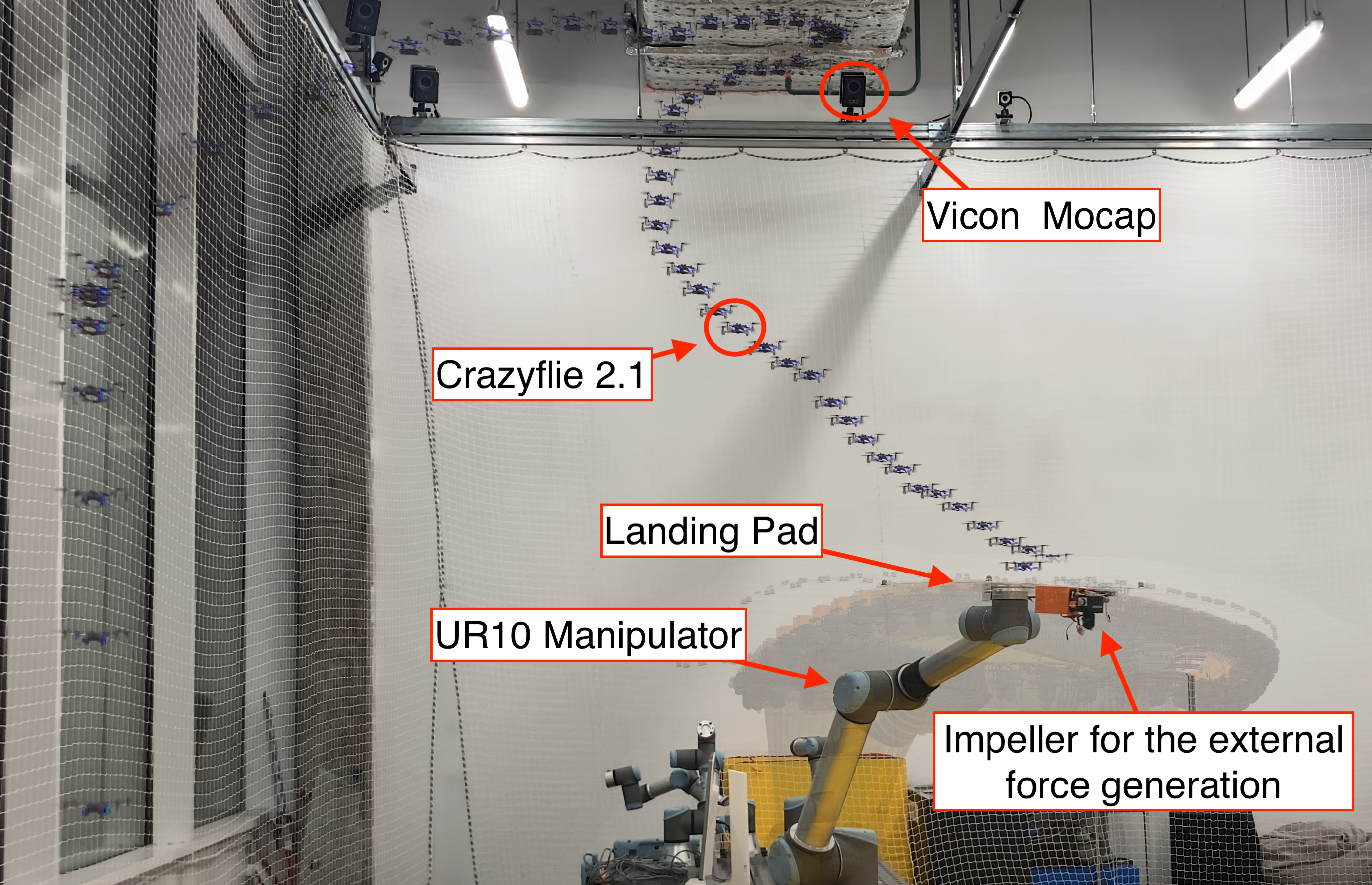}
 \caption{Composite image of the Lander.AI drone agent landing on UR10 robotic arm equipped with landing pad and air disturbance impeller.}
 \label{fig:test}
\end{figure}
Achieving accurate landing on a moving platform in the presence of the wind effect is not just a technical accomplishment, it represents a significant advancement in the operational capabilities of drones for real-world applications.

Deep Reinforcement Learning (DRL) has emerged as a promising approach. DRL, a specialized branch of machine learning, involves training algorithms via a trial-and-error approach, enabling them to determine optimal actions in complex and unpredictable environments. This methodology is exceptionally suited for autonomous drone landing tasks, where drones must make real-time decisions based on immediate environmental feedback.

This research introduces the Lander.AI agent (Fig. \ref{fig:test}), utilizing Deep Reinforcement Learning for effective landing and motion planning under unpredictable conditions, including sudden wind changes and moving platform velocities. We utilize indoor localization system (Vicon) to test the agent in real world conditions, it effectively guide the drone to land on moving platform with uncertainties.

The agent's performance, trained in a simulated environment and tested on Crazyflie 2.1 drones, is benchmarked against a baseline PID controller with an Extended Kalman filter, demonstrating Lander.AI's adaptability. This work significantly advances drone autonomy and safety, potentially transforming their deployment in dynamic scenarios.

Our Lander.AI agent revolutionizes drone landing with bio-inspired learning, intuitively handling external forces like wind without exact force data. Its training transcends specific drone specs, ensuring broad adaptability through domain randomization. Lander.AI deciphers indirect flight dynamics to seamlessly counter environmental challenges, mirroring birds' natural flight adaptability. This approach significantly enhances drone flexibility across diverse scenarios without custom modifications.

\section{Related Works}
The realm of autonomous UAV landing has evolved significantly, driven by advancements in vision-based techniques, DRL, and innovative landing strategies. The corpus of research spans a broad spectrum, from precision landings on static platforms to the complex dynamics of moving platform engagements, underpinned by a keen focus on environmental adaptability and operational robustness.

Vision based approach is one classical approach. Authors of \cite{b1}, \cite{b2}, \cite{b21} and \cite{b8} highlight the fusion of vision-based systems with DRL to enhance UAV landing capabilities. These demonstrate the potential of real-time visual inputs and DRL for precise landings. However, they also reveal the sensitivity of vision-based systems to environmental variables, such as lighting and weather conditions, which can impact system performance, particularly in dynamic landing scenarios.

Research efforts, particularly those early in the field, have often focused on static platform landings, Authors of \cite{b1}, \cite{b3}, and \cite{b28} showcasing significant advancements in this area. These works, however, primarily concentrate on the precision aspect without extensively tackling the unpredictable dynamics associated with moving platforms. Several recent works \cite{b4}, \cite{b11}, delve into the nuanced challenges of landing on moving platforms. These studies introduce adaptive algorithms and strategies to address the inherent unpredictability of moving targets but lack a comprehensive treatment of external environmental factors like wind forces. 

The impact of wind forces and changing velocities is a critical consideration for autonomous UAV landings. While studies such as \cite{b26} directly address wind turbulence, the integration of these environmental factors with the challenges of moving platform landings is less frequently explored in depth. Authors of \cite{b5} worked on moving platform drone landing using reinforcement learning combined with PID control with changing UGV velocities. There is a notable gap in the literature regarding the holistic consideration of wind forces in conjunction with the dynamics of moving platforms.

Authors of the research \cite{b3}, \cite{b6}, \cite{b16} underscore the advancements in landing precision and the adaptability of UAVs facilitated by DRL. These contributions highlight the potential of DRL in enhancing UAV responsiveness to complex dynamics. However, the extensive computational demands and the need for substantial training data are recurrent challenges that can limit the scalability and real-world applicability of these systems.

Research works in \cite{b9}, \cite{b14}, \cite{b22}, and innovative approaches presented in \cite{b12} and \cite{b13}, explore the boundaries of DRL applications in UAV operations, from high-speed maneuvers to novel landing techniques like tethered perching. These studies open new avenues for UAV landing technologies but also underscore the need for adaptable and generalizable models capable of navigating the diverse requirements of real-world applications.

The broader landscape, including research presented in \cite{b10}, \cite{b17}, \cite{b18}, and etc., emphasizes the critical role of environmental adaptation. The challenge of operating within complex and variable real-world conditions, including wind turbulence and dynamic platform movements, is a recurring theme. These studies point to a need for models that can robustly account for a wide range of external factors, highlighting a gap in current research regarding the comprehensive integration of these elements.

Collective body of research on autonomous UAV landing using DRL has made significant strides in addressing precision, adaptability, and the development of novel landing strategies. However, a more integrated approach that holistically considers static and moving platforms, alongside environmental challenges such as wind forces and velocity changes, remains an area ripe for further exploration. We Address these interconnected challenges and propose Lander.AI agent to do advancing the operational safety, reliability, and efficiency of UAV landings in the face of the complex and unpredictable conduitions of real-world environments.

\section{Methodology}

This section outlines the approach employed to develop and validate the Lander.AI agent, focusing on autonomous drone landing on dynamically moving platforms under varying environmental conditions using Deep Reinforcement Learning framework.

\subsection{Simulation Environment Setup}
\begin{figure}[h]
 \centering
 \includegraphics[width=0.4\textwidth]{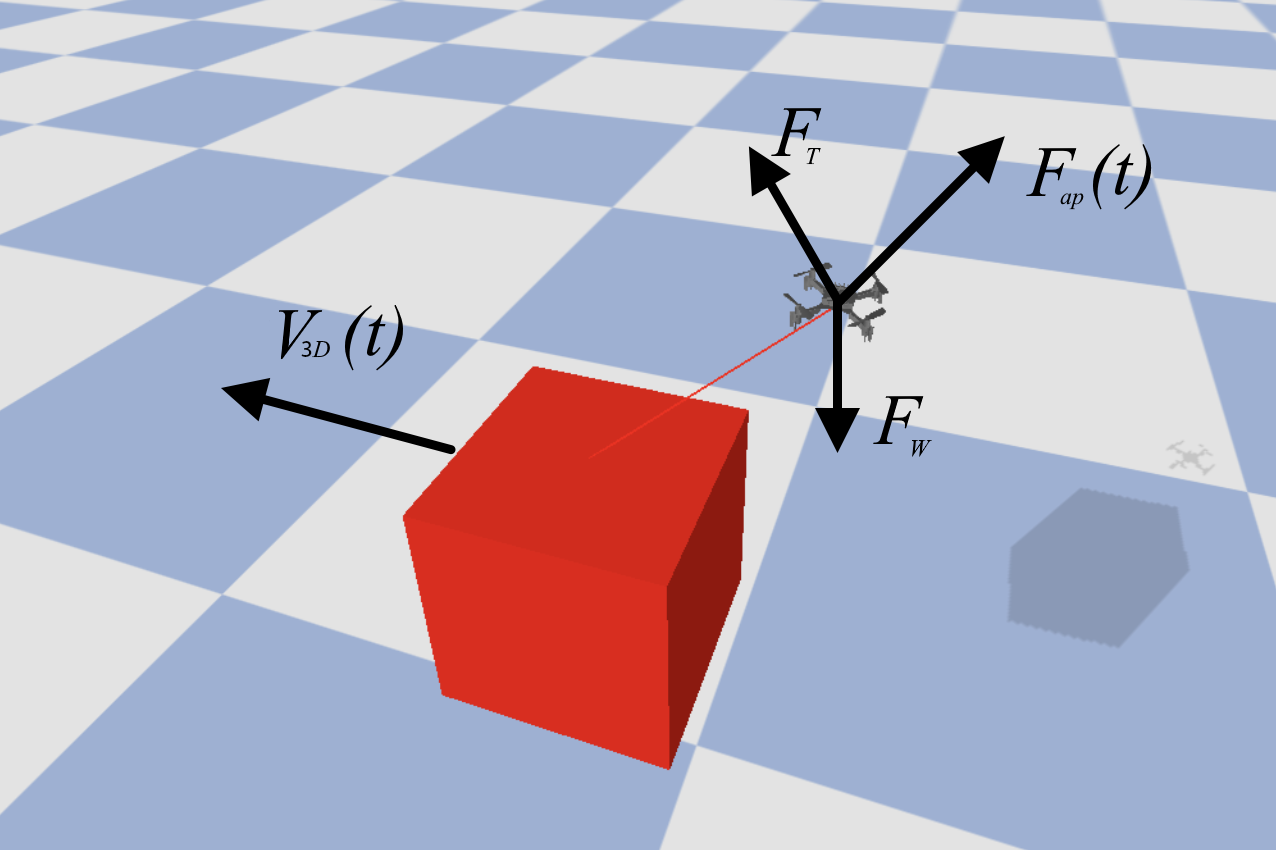}
 \caption{Gym-pybullet Simulation Environment Setup.}
 \label{fig:land}
\end{figure}

Our simulation environment is developed using the Gym framework and PyBullet physics, featuring configurations that emulate the Crazyflie 2.x drone for realistic aerodynamic simulations, uses gym-pybullet-drones \cite{b15}. Central to our setup is a 0.5 m cubical platform, depicted in Figure \ref{fig:land}, which moves in the XYZ space with velocities ranging from -0.46 to 0.46 m/s, thus introducing a dynamic challenge for precision landings.

The drone's parameters include linear velocities from -3 to 3 m/s in XY and -2 to 2 m/s in Z, and rotation angles from $-\pi$ to $\pi$ radians, necessitating adaptive flight strategies for effective landings.

To enhance the Lander.AI agent's training and adaptability to external disturbances, we introduced a probabilistic framework for applying random external forces:

\begin{equation}
F =
\begin{cases}
F_{ap}(t) & \text{if } p(e) < 0.2 \\
0 & \text{if } p(e) \geq 0.2
\end{cases} ,
\end{equation}

\begin{equation}
F_{ap}(t) =
\begin{cases}
\text{sgn}(f(t,\xi)) \times |\text{f}(t, \xi)| & \text{if } p(s) < 0.2 \\
0 & \text{if } p(s) \geq 0.2 
\end{cases} ,
\end{equation}
where $p(e)$ is the probability at which the force will be applied during the episode, $p(s)$ is the probability that the force will be applied at the current step of the “windy" episode, $F$ is the vector of an external force based on the binary indicator $f_{i}$. The force direction is selected randomly with $x$, $y$, and $z$ components in world coordinate frame being in range of -0.005 to 0.005, simulating realistic environmental disturbances like wind. This method aims to increase the agent's resilience and performance under varied and unpredictable conditions typical in real-world operational scenarios.

\subsection{Deep Reinforcement Learning Framework}

\subsubsection{Lander.AI Agent Architecture}
The Lander.AI agent employs a neural network tailored for drone landing, utilizing observations as follows: 
\begin{equation}
 \vec{o}_t = \left[ \vec{\theta}, \vec{v}, \vec{\omega}, \vec{d}, \Delta \vec{v} \right],
\end{equation}
which corresponds to the attitude of the drone (roll, pitch, yaw), linear velocity, angular velocity, relative landing pad positions and relative velocities of the landing pad. These inputs are first clipped and normalized to a range of -1 to 1, ensuring optimal neural network performance.

\begin{figure}[h]
 \centering
 \includegraphics[width=0.4\textwidth]{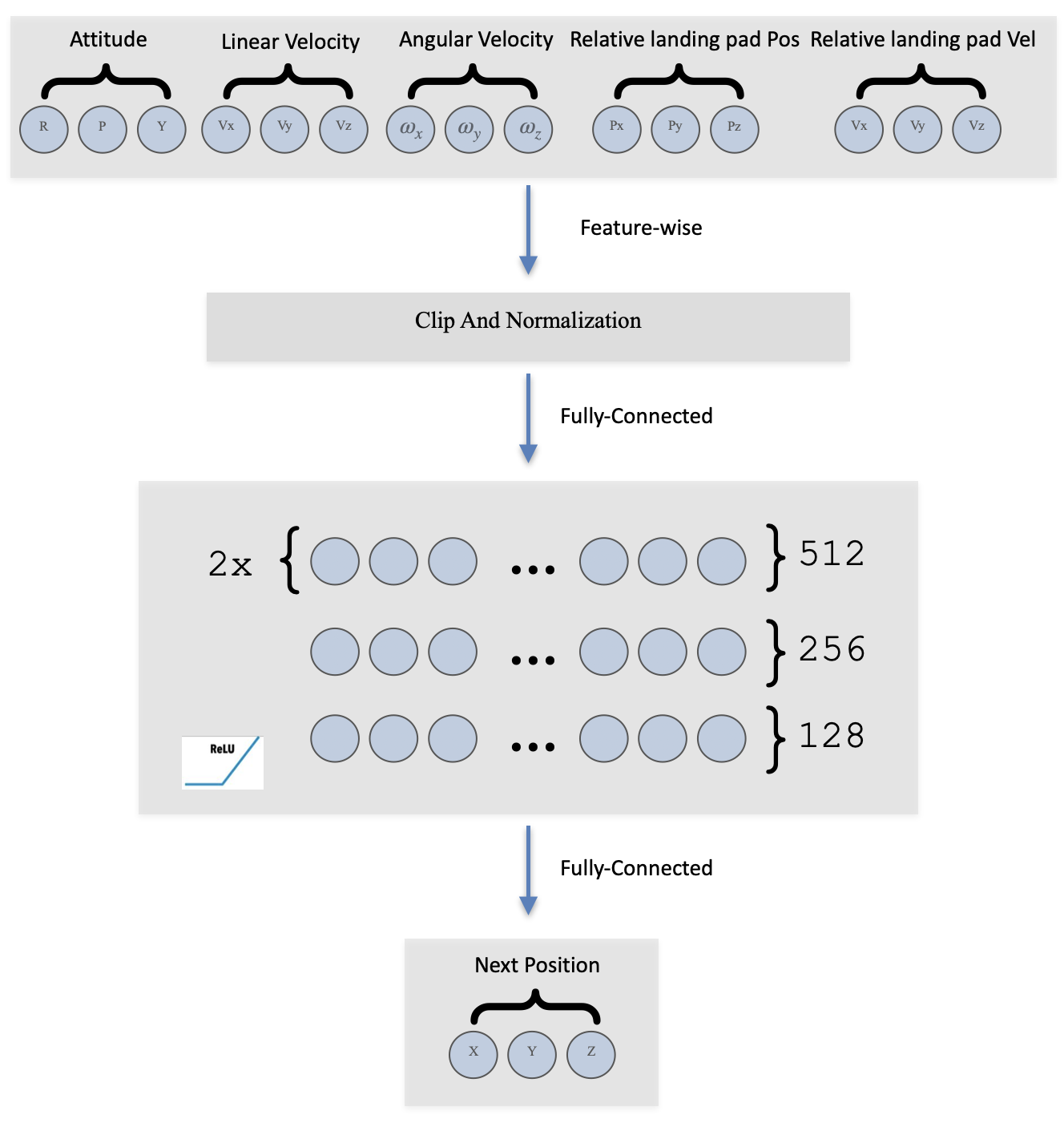}
 \caption{Architecture of Deep Reinforcement Learning Model Illustrating Inputs, Hidden Layers, and Action Mechanisms.}
 \label{fig:rl}
\end{figure}

Figure \ref{fig:rl} represents the neural network architecture behind our deep reinforcement learning approach.


\begin{equation}
\resizebox{0.9\linewidth}{!}{$\text{ReLU}(\text{FC}{512 \times 2}(\hat{\vec{o}}t)) \rightarrow \text{ReLU}(\text{FC}{256}) \rightarrow \text{ReLU}(\text{FC}{128})$} 
\end{equation}
where FC are the fully-connected layers with ReLU activation functions, arranged in dimensions of 512x2, 256, and 128. This setup processes the standardized inputs to determine the drone's precise adjustments for landing. The output layer, with three neurons, generates PID control signals dictating position changes in the range of -1, 0, and 1. 

\begin{equation}
\Delta \vec{p}_t = 0.1 \times \vec{c}_t , 
\end{equation}
where $\vec{c}_t$ is the control signal for position change $\Delta \vec{p}_t$. These adjustments are applied to the current drone pose, with a 0.1 factor, guiding the drone towards an accurate landing.

\subsubsection{Reward Function}
The agent's reward function is crafted to enhance precision and adaptability in landing. It is structured as follows:

\begin{equation}
\resizebox{0.9\linewidth}{!}{$
\begin{split}
\text{Reward} = 
 \begin{cases}
 \tanh(\gamma), & \text{if } d_{target} > 2 \\
 \tanh(\alpha \times (d_{target} - R)), & \text{if } d_{target} \subseteq (0.1, 2) \\
 \tanh(-U - \beta + \Delta), & \text{if } d_{target} < 0.1\\
 \tanh(-U + \Delta), & \text{Otherwise}
 \end{cases}
\end{split}$}
\end{equation}

where $\gamma$ is the penalty reward for moving far away from the target, $d_{target}$ is the distance between drone and the target landing pad, $\alpha$ is the reward scaling factor for proximity to the target, $R$ is the current distance to the target. $U = U_{attractive} + U_{repulsive}$ combines attractive and repulsive potentials. $\beta$ adjusts for edge proximity penalties and below the landing pad altitude. $\Delta$ discourages excessive speed allowing descending relative velocity while approaching landing pad .


The attractive and repulsive potentials are defined as:
\begin{equation}
\resizebox{0.9\linewidth}{!}{$
U_{repulsive} = 
\begin{cases} 
 \frac{1}{2} \times \eta \times \left( \frac{1}{\sigma} - \frac{1}{Q_{max}} \right)^2, & \text{if } \sigma < Q_{max} \\
 0, & \text{Otherwise}
\end{cases}$}
\end{equation}

\begin{equation}
U_{attractive} = \frac{1}{2} \times \zeta \cdot R^2
\end{equation}
where $\eta$ is the strength of the repulsive potential, $\sigma$ is the distance to the nearest obstacle, $Q_{max}$ is the maximum effective distance of the repulsive potential, $\zeta$ is the strength of the attractive potential, $R$ is the current distance to the target.


This reward function dynamically balances the Lander.AI agent's objectives, guiding it towards successful landings while avoiding hazards and ensuring smooth descent trajectories.
\begin{figure}[h]
 \centering
 \begin{minipage}{0.45\textwidth}
 \centering
 \includegraphics[width=\textwidth]{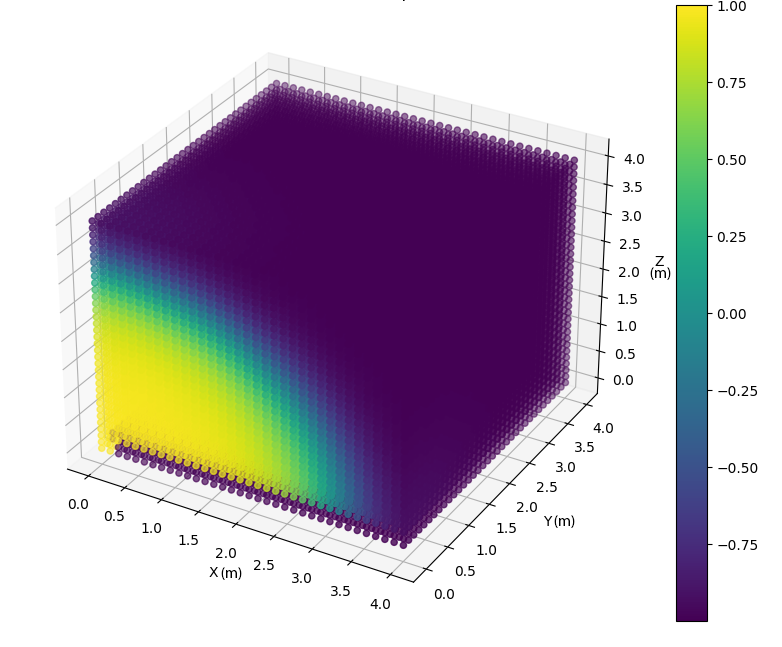}
 \caption{Origin View of Lander.AI's Reward Function Emphasizing Safety and Behavior.}
 \label{fig:reward1}
 \end{minipage}\hfill 
 \begin{minipage}{0.45\textwidth}
 \centering
 \includegraphics[width=\textwidth]{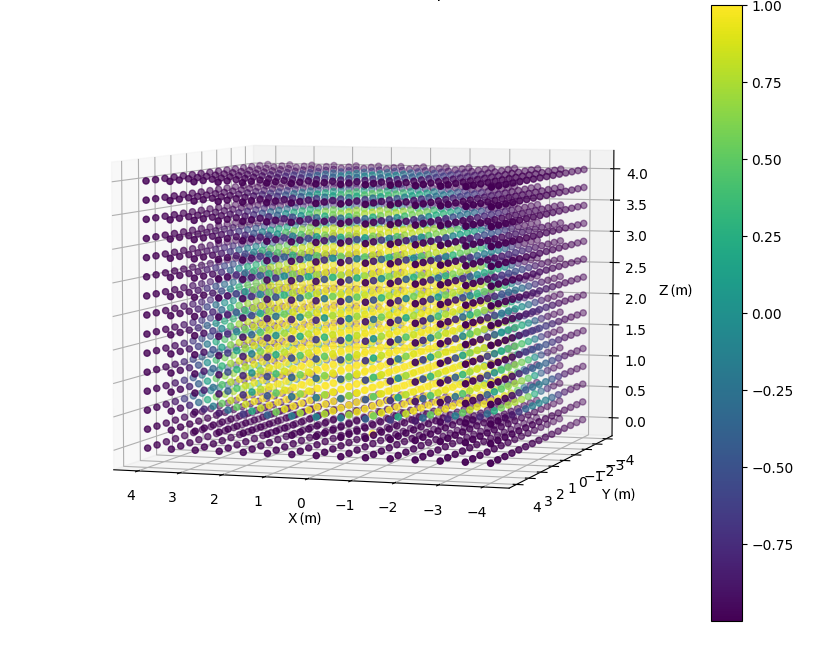}
 \caption{Lateral View of Lander.AI's Reward Function Showcasing Safety and Behavioral Rewards.}
 \label{fig:reward2}
 \end{minipage}
\end{figure}

The reward function of the Lander.AI agent, integral to our methodology, is constructed using a potential field approach and is depicted in 3D space for comprehensive visualization. Currently, the function employs an attractive potential field to guide the drone towards its landing target, with a reward gradient that enhance precision by offering higher rewards closer to the target zone, as illustrated in Figures \ref{fig:reward1} and \ref{fig:reward2}. This gradient is apparent from the origin view and the side view, with the color transition from yellow to dark purple indicating the shift from optimal to less desirable states.

In addition to the attractive potential field, a crucial safety mechanism is embedded within the reward structure, which penalizes the drone from operating below a predefined altitude relative to the landing pad. This safety reward ensures that the drone maintains a safe approach trajectory and does not fly at an altitude that would be considered hazardous or below the landing platform's level.

Looking ahead, the reward function is poised for expansion to include a repulsive potential field. This future development aims to further sophisticate the agent's navigational capabilities by introducing negative rewards for approaching obstacles, thereby preventing collisions and reinforcing safe flight paths in complex environments.
\subsubsection{Training Protocol}

The Lander.AI agent's training protocol is meticulously structured to ensure an effective learning progression. Using the stable-baselines3 Twin Delayed DDPG (TD3) algorithm \cite{b27}, the agent undergoes a rigorous training regimen designed for complex and continuous control tasks. The choice of TD3 is motivated by its demonstrated ability to converge more rapidly compared to alternative algorithms, particularly in environments with continuous and complex behaviors.

The policy employed is the MlpPolicy with an initial learning rate of 0.0001. The agent was initially trained over 5 million steps, with each episode capped at 20 seconds to allow the agent to acquire the main behavioral patterns necessary for landing. To further refine the agent's capabilities, including additional safety maneuvers and adaptability skills, the model underwent retraining up to 35 million steps. This extended training involved multiple iterations of fine-tuning in both simulated environments and real-world testing scenarios, enhancing performance in dynamic 3D spaces where the landing pad presents complex patterns and sudden directional changes.

The training leverages a buffer size of 1,000,000 (1e6), with learning commencing after 100 steps and a batch size of 100. The MlpPolicy parameters are set to use ReLU activation functions, a FlattenExtractor for feature extraction, and image normalization is enabled. The optimizer of choice is Adam, known for its efficiency in handling sparse gradients on noisy problems.

\begin{figure}[h]
 \centering
 \begin{minipage}{0.5\textwidth}
 \centering
 \includegraphics[width=\textwidth]{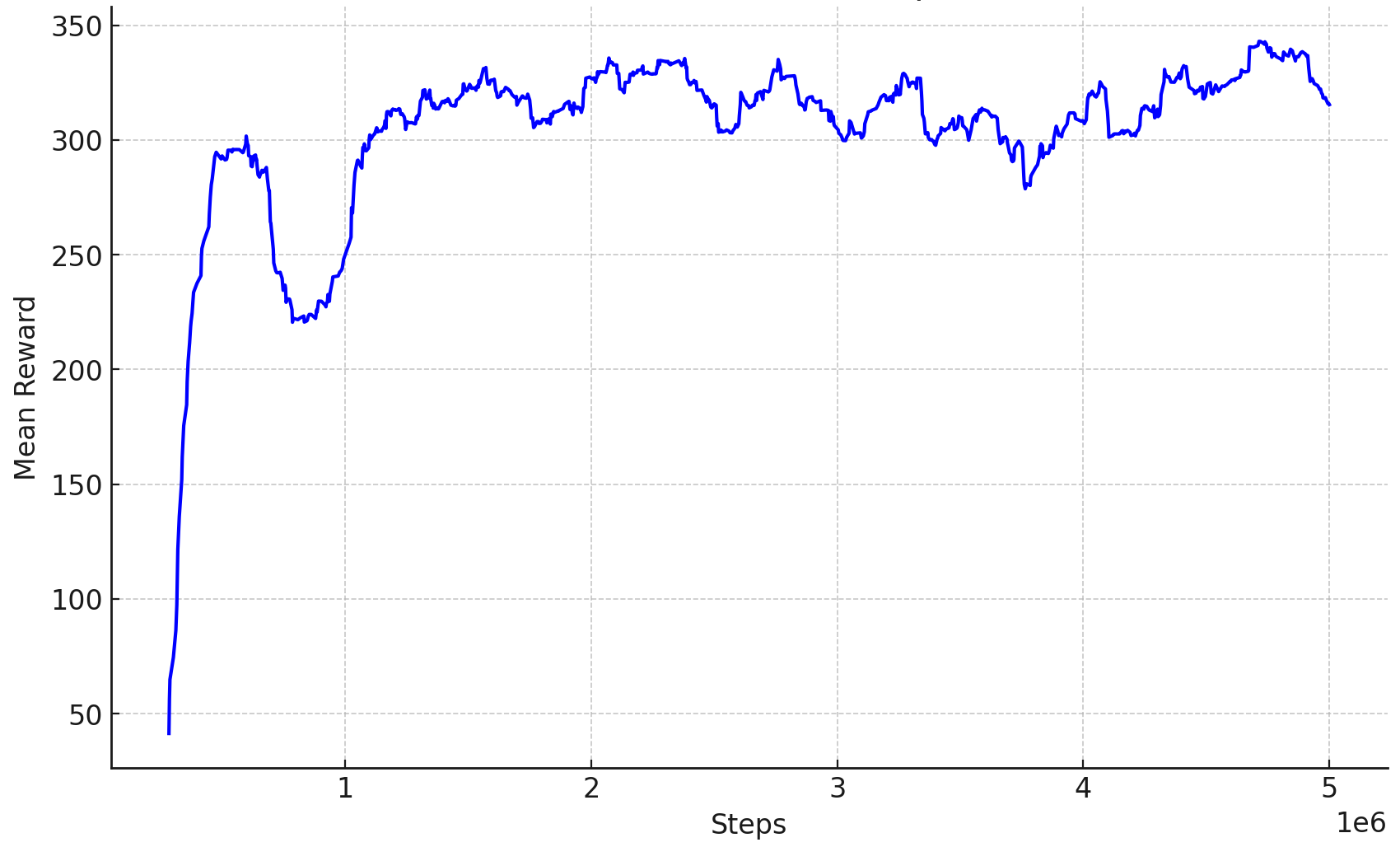}
 \caption{Mean Reward vs Training Steps, showcasing learning progress.}
 \label{fig:rewardstep}
 \end{minipage}\hfill
\\
 \begin{minipage}{0.5\textwidth}
 \centering
 \includegraphics[width=\textwidth]{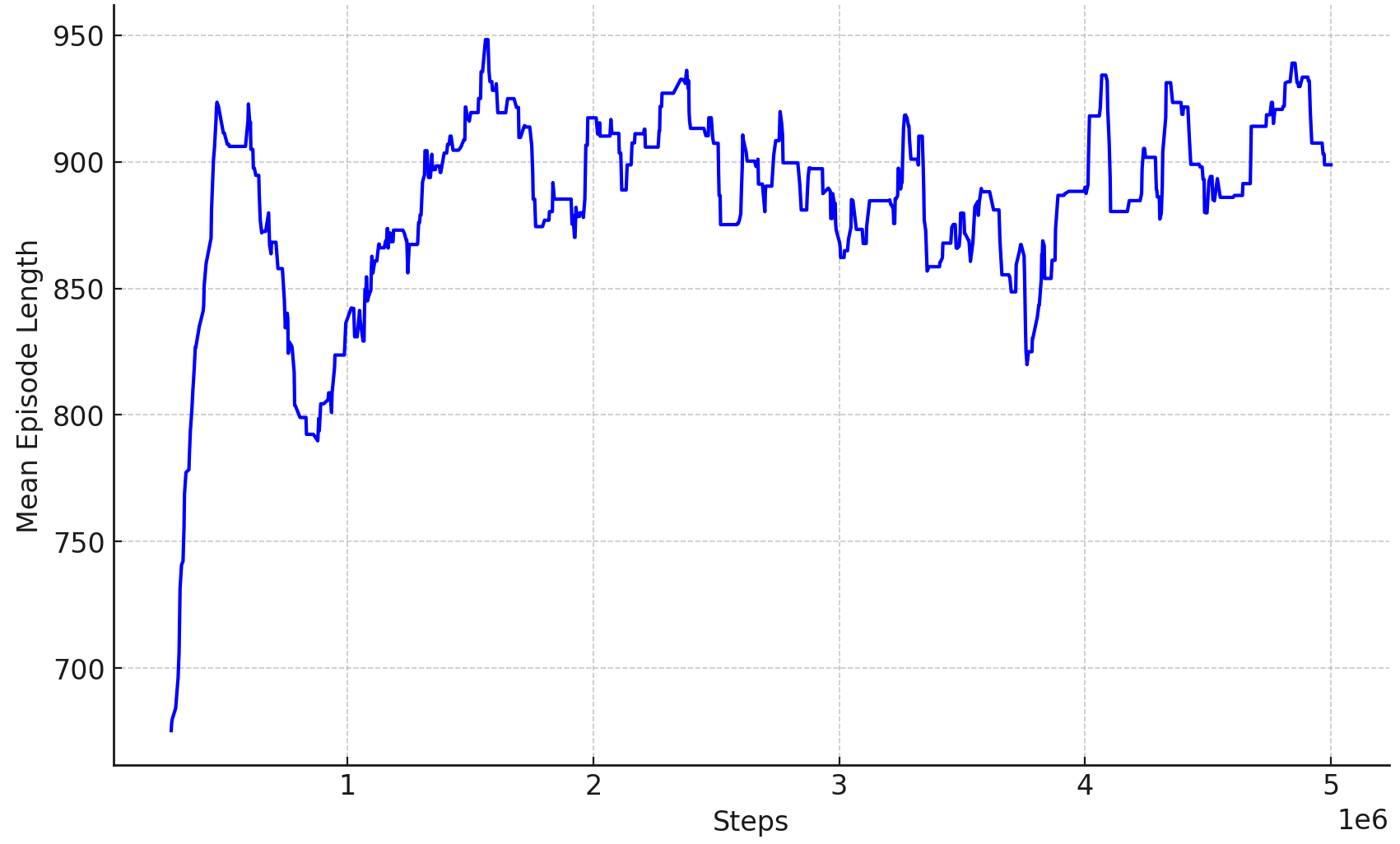}
 \caption{Mean Episode Length vs Training Steps, indicating agent endurance.}
 \label{fig:epstep}
 \end{minipage}
\end{figure}

This training protocol culminates in agent navigating in 3D spaces with dynamic landing platforms, demonstrating quick adaptation to unforeseen environmental changes and complex landing trajectories.
Figures \ref{fig:rewardstep} and \ref{fig:epstep} capture the agent's learning progress, with an increasing mean reward and episode length over training steps showcasing the agent's enhanced reward optimization and sustained performance throughout the learning phase.

\subsection{Real-World Validation Setup}
\subsubsection{Indoor Localization System (Vicon)}

For real-world testing, we employed a Vicon motion capture system to provide high-precision localization of both drones and platforms. This system delivers positional data at a rate of 100Hz, which is critical for extracting accurate observations necessary for the Lander.AI agent's operation. The observations are then fed into the agent to inform its decision-making process. The Vicon system's VRPN (Virtual-Reality Peripheral Network) positioning type ensures a robust and precise tracking capability, vital for the successful deployment and testing of our autonomous navigation algorithms in a controlled indoor environment.

\subsubsection{Crazyflie Drones}

The empirical tests were conducted using Crazyflie2.1 drones, which are equipped with onboard default PID controllers for low-level flight control. The system was integrated with ROS2, which facilitated the issuance of high-level position commands. Communication with the drones was achieved through a Crazyradio 2.4 GHz RF transmitter, operating at a frequency of 100Hz to ensure real-time responsiveness and precise maneuvering during flight tests.

\subsection{Baseline Comparison}
\subsubsection{Crazyflie On-board PID Controller with Extended Kalman Filter}

The baseline for our comparative analysis incorporates a PID control system enhanced with an Extended Kalman Filter (EKF) for precise tracking of moving landing platforms. The EKF implementation is tailored to predict and update the platform's position and velocity, ensuring accurate tracking under dynamic conditions. The core of the EKF is defined and initialized with the platform's initial state, covariance, and the variances associated with the process and measurements.

The state transition matrix $\mathbf{A}$ and observation matrix $\mathbf{H}$ are constructed as follows:
\[
\mathbf{A} = \begin{bmatrix}
 1 & 0 & 0 & 1 & 0 & 0 \\
 0 & 1 & 0 & 0 & 1 & 0 \\
 0 & 0 & 1 & 0 & 0 & 1 \\
 0 & 0 & 0 & 1 & 0 & 0 \\
 0 & 0 & 0 & 0 & 1 & 0 \\
 0 & 0 & 0 & 0 & 0 & 1
\end{bmatrix}, \quad
\mathbf{H} = \begin{bmatrix}
 1 & 0 & 0 & 0 & 0 & 0 \\
 0 & 1 & 0 & 0 & 0 & 0 \\
 0 & 0 & 1 & 0 & 0 & 0
\end{bmatrix}
\]

The `predict' method advances the state estimation based on the motion model, while the `update' method refines this estimation with incoming measurements, employing the Kalman Gain to minimize the estimation error. This EKF framework serves as a robust baseline, facilitating a comprehensive evaluation of the Lander.AI agent's performance in tracking and landing on moving platforms.

\section{Experiments}
Our experiments are designed to explore key aspects of autonomous drone landing: (i) Comparing the Lander.AI agent's landing accuracy and consistency against traditional control methods on moving platforms. (ii) Assessing the Lander.AI agent's resilience to environmental disturbances and dynamic platform behaviors. (iii) Evaluating the agent's versatility across varied and complex landing scenarios. (iv) Validating the agent's simulation-trained strategies in real-world settings.

\subsubsection{Experimental Design}
Our real-world testing framework was meticulously designed to validate the Lander.AI agent under various conditions. Utilizing a UR10 robotic arm, we mounted a $0.5 \times 0.5 \times 0.003$ meter acrylic landing pad on its TCP, ensuring precise and controlled movements. To simulate air disturbances, an impeller powered by a 12V battery through an Arduino Uno was embedded in the landing pad.

The experimental setup was divided into four distinct scenarios to comprehensively evaluate landing performance which is 10 test cases per scenarios, overall 80 test cases:

\begin{enumerate}
 \item \textbf{Static Point Landing (SPL)}: Testing the agent's ability to land on a stationary platform.
 \item \textbf{Linear Moving Point Landing (LMPL)}: Assessing landings on a platform moving linearly with sudden directional changes.
 \item \textbf{Curved Moving Point Landing (CMPL)}: Evaluating landings on a platform following a curved trajectory with directional shifts.
 \item \textbf{Complex Trajectory Landing (CTL)}: The Lander.AI agent's adaptability is further tested through challenging landings on dynamically moving platforms in three-dimensional space, amidst wind disturbances generated by impellers mounted on the landing pads..
\end{enumerate}
\begin{figure*}
 \centering
\begin{subfigure}[b]{0.24\textwidth}
\includegraphics[width=\textwidth]{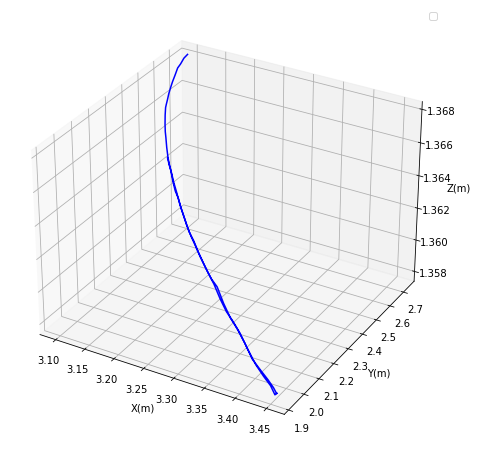}
\caption{}
\end{subfigure}
\begin{subfigure}[b]{0.24\textwidth}
\includegraphics[width=\textwidth]{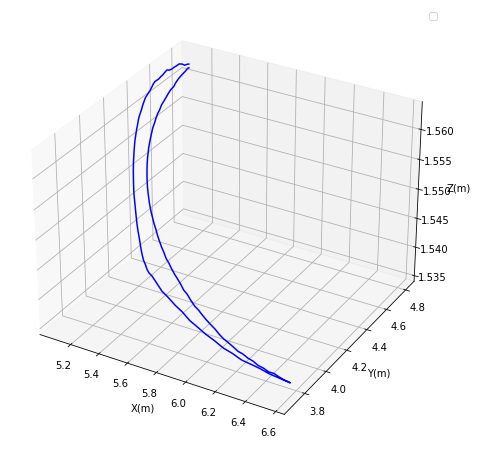}
\caption{}
\end{subfigure}
\begin{subfigure}[b]{0.24\textwidth}
\includegraphics[width=\textwidth]{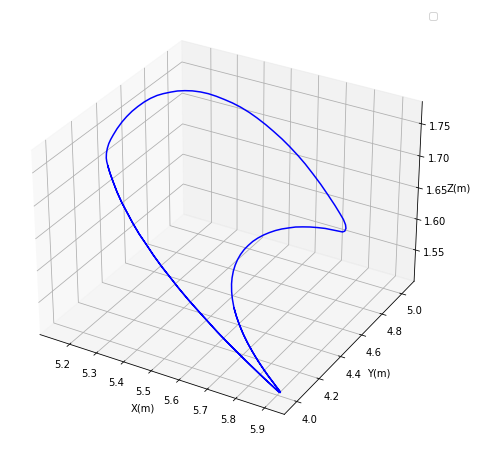}
\caption{}
\end{subfigure}
\caption{Trajectories of the moving landing pad (a) linear, (b) curved, and (c) complex 3D motion showing our experiment setup.}
\label{fig:landingpads1}
 \end{figure*}

\begin{figure*}
\centering
\begin{subfigure}[b]{0.24\textwidth}
\includegraphics[width=\textwidth]{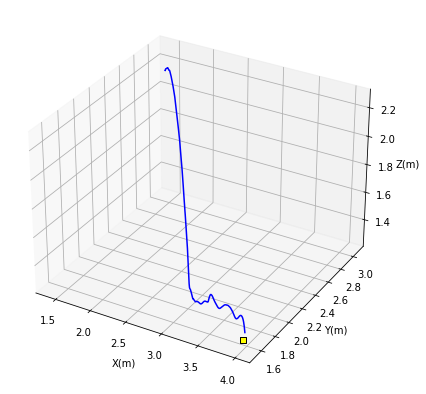}
\caption{}
\end{subfigure}
\begin{subfigure}[b]{0.24\textwidth}
\includegraphics[width=\textwidth]{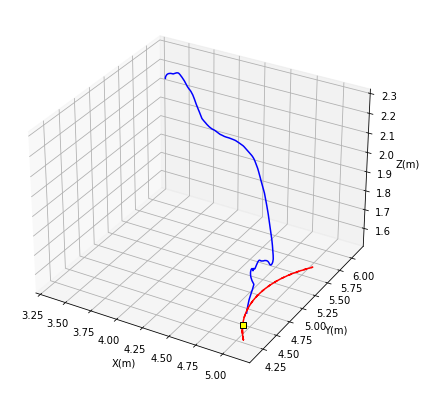}
\caption{}
\end{subfigure}
\begin{subfigure}[b]{0.24\textwidth}
\includegraphics[width=\textwidth]{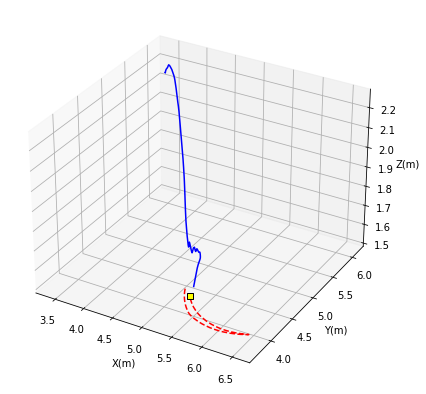}
\caption{}
\end{subfigure}
 \begin{subfigure}[b]{0.24\textwidth}
 \includegraphics[width=\textwidth]{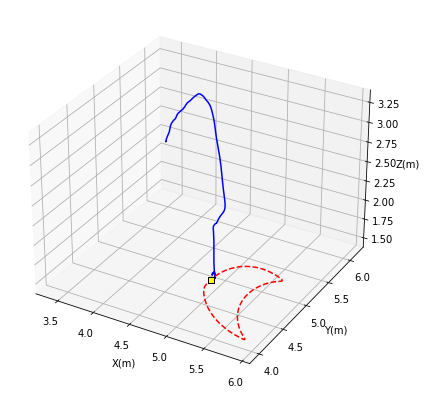}
 \caption{}
 \end{subfigure}

\caption{Trajectories of the moving landing pad (red line) and drone (blue line) in (a) fixed point, (b) linear, (c) curved and (d) complex 3D motion showing agent adaptability to different scenarios.}
 \label{fig:traj1}
 \end{figure*}

Each scenario was tested in 10 trials, comparing our agent's performance against a baseline established using an Extended Kalman Filter (EKF) with the onboard Crazyflie PID controller. To further test the agent's adaptability, scenarios 1 and 2 were also conducted with air disturbances generated by the impeller, exclusively for the Lander.AI agent.The experimental setup for moving landing pad trajectories, including linear, curve, and complex 3D motions, is depicted in Figure ~\ref{fig:landingpads1}.

\subsubsection{Performance Metrics}
\begin{figure}[h]
 \centering
 \includegraphics[width=0.45\textwidth]{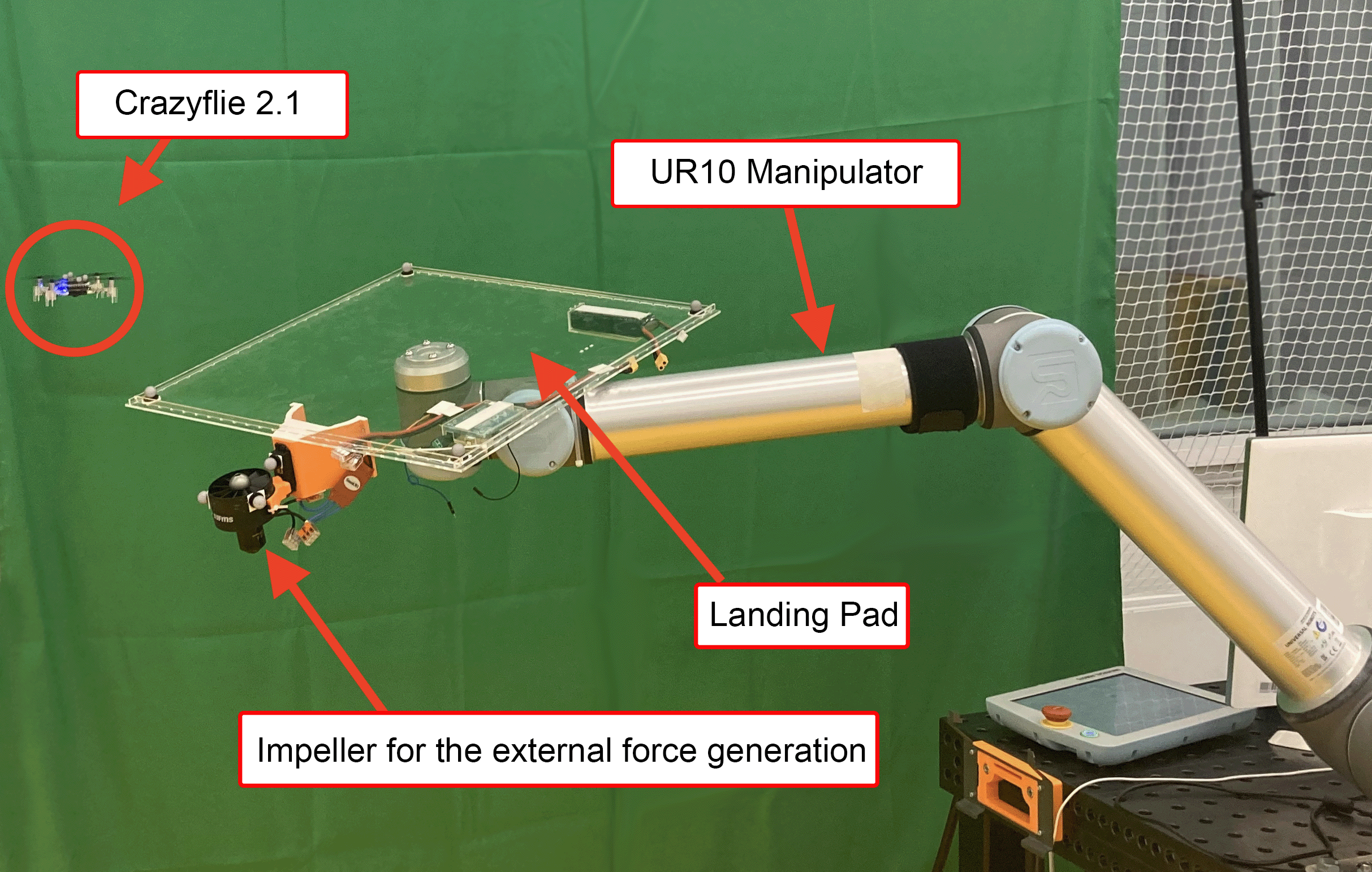}
 \caption{Experimental setup with drone landing on a moving platform in the presence of external force.}
 \label{fig:reallanding}
\end{figure}

To evaluate the algorithm effectiveness against the baseline controller, we employ a comprehensive set of performance metrics:

\begin{enumerate}
 \item \textbf{Landing Success Rate}: The proportion of successful landings on dynamic targets under varied conditions is quantified to gauge both consistency and reliability across trials.
 \begin{table}[h!]
\caption{Comparison of Landing Success Rates: Lander.AI \\ Agent vs. EKF with PID Controller}
 \centering
 \begin{tabular}{|l|l|l|}
 \hline
Test Case & Lander.AI Agent & EKF with PID \\ \hline
 SPL & 100\% & 80\% \\ \hline
 LMPL & 100\% & 20\% \\ \hline
 CMPLanding & 100\%& 40\% \\ \hline
 CTL & 60\% & 10\% \\ \hline
 \end{tabular}
 \label{table:landing_success_rate}
 \end{table}
 
 Table \ref{table:landing_success_rate} compares the Lander.AI agent's landing success rates with those of a traditional EKF-PID control system across four scenarios: Static, Linear, Curved, and Complex Trajectory Landings. Expressed as percentages, the results demonstrate the Lander.AI agent's enhanced adaptability and precision, significantly outperforming the traditional system in both static and dynamic contexts.
 
 \item \textbf{Landing Precision}: This metric evaluates the drone's accuracy in reaching a specific point on the moving target, with precision gauged by the mean distance from the target point across multiple attempts.
 
 \begin{table}[h]
\caption{Comparative Analysis of Landing Precision: Lander.AI Agent vs EKF with PID}
\centering
\begin{tabular}{|l|c|c|c|c|} 
 \hline
Test Case & \multicolumn{2}{c|}{Lander.AI Agent} & \multicolumn{2}{c|}{EKF with PID} \\ 
 \hline
 {} & Mean & STD & Mean & STD \\ 
 \hline
 SPL & 0.0524m & 0.0020m & 0.1365m & 0.0377m \\ 
 \hline
 LMPL & 0.0650m & 0.0214m & 0.1237m & 0.0634m \\ 
 \hline
 CMPL & 0.0945m & 0.0241m & 0.1357m & 0.0741m \\ 
 \hline
 CTL & 0.1377m & 0.0537m & 0.2022m & 0.0516m \\ 
 \hline
\end{tabular}
\label{table:landing_precision}
\end{table}
 Table \ref{table:landing_precision} presents a detailed comparison of landing precision between the developed system and a control system employing an Extended Kalman Filter (EKF) with a PID controller across four experimental setups: Static Point, Linear Moving Point, Curved Moving Point, and Complex Trajectory Landings. Precision is quantified by the mean distance from the target landing point, with associated standard deviations (STD) indicating variability across trials. The Lander.AI agent consistently demonstrates superior precision with lower mean distances and tighter standard deviations compared to the EKF-PID system. This trend is evident across all scenarios, highlighting the Lander.AI agent's enhanced capability to accurately navigate and land, especially in dynamically challenging environments.
 
 \item \textbf{Complexity of Scenarios and Recovery from Perturbations}: The agent's performance is tested against a spectrum of complex situations, including unpredictable target movements and challenging environmental conditions, as well as its capacity to stabilize and land following disturbances such as wind gusts or abrupt target motion changes, to ascertain its versatility, real-world applicability, adaptability, and resilience. Summary of correlation between drone velocity vs landing pad velocity statistics and comments are provided in Table \ref{tab:summary}.

 \begin{table}[h!]
 \caption{Summary for Complexity of Scenarios and Recovery from Perturbations}
 \centering
 \begin{tabular}{|l|c|c|c|c|c|}
 \hline
 Correlation & SPL & LMPL & CMPL & CTL \\ \hline
 Mean & 0.1012 & 0.5822 & 0.5028&0.2340 \\ \hline
 Median & 0.1020 & 0.6055 &0.4947 &0.2153 \\ \hline
 STD & 0.0674 & 0.1174 &0.2320 & -0.2399 \\ \hline
 Min & -0.0067 & 0.3561 & -0.0468 & -0.1106\\ \hline
 Max & 0.1829 & 0.7407 & 0.7703 & 0.5369 \\ \hline
 \end{tabular}
 \label{tab:summary}
\end{table}

\end{enumerate}
The real world experimental setup with drone landing on a moving
platform in the presence of external force is demonstrated in Figure~\ref{fig:reallanding}.

The results of the experiment display a correlation between drone and landing pad velocities across diverse test cases assessing the Lander.AI agent's adaptability.

\begin{figure}[ht]
 \centering
 \includegraphics[width=0.4\textwidth]{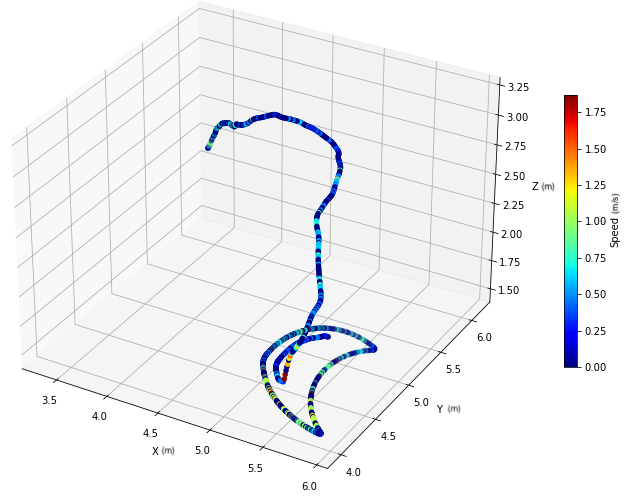}
 \caption{Velocity changes of Drones and Moving pad explaining the complexity}
 \label{fig:velocity}
\end{figure}

Notably, higher mean correlations suggest better synchronization. Strong adaptability is evident in linear moving point landings, while variability in correlation coefficients reflects resilience levels. Standard deviation and correlation ranges provide insights into consistency and robustness. These metrics offer quantitative assessments of the Lander.AI agent's performance in varied environments and perturbations.
The velocity changes and their complexities are illustrated in Figure~\ref{fig:velocity}. 

The velocity analysis reveals that the highest velocity snap of 1.75 m/s happens during the final stage of the landing where the agent adapts to the ground effect and external force simultaneously. However, the change in trajectory at this point is compensated by the DRL agent, showcasing a successful landing performance. The adaptability of the developed agent to various scenarios is illustrated through drone landing trajectories in Figure~\ref{fig:traj1}, covering fixed point, linear, curve, and complex 3D motions.

\section{CONCLUSION AND FUTURE WORK}

In conclusion, this study advances autonomous drone navigation by introducing the Adaptive Landing Behavior Agent, which excels in dynamic landings and outperforms traditional methods like the EKF with PID controllers. Lander.AI achieved a 100\% success rate in static and linear scenarios, and 60\% in complex trajectories, significantly better than the EKF with PID's results. Notably, Lander.AI demonstrated superior landing precision, with a mean distance to the target of 0.08m, outperforming the EKF with PID setup's 0.14m, and achieving a best-case distance of just 0.05m.

Lander.AI's adaptability, especially in linear scenarios with a mean correlation of 0.5822 with moving targets, highlights the effectiveness of deep reinforcement learning in aerodynamic challenges. These results underscore the system potential in enhancing drone landing precision and reliability for critical applications like emergency response and logistics.

Future work will focus on refining Lander.AI through deep reinforcement learning to autonomously adapt to external disturbances without explicit pre-training, akin to birds' instinctive flight adjustments. This approach aims to enable Lander.AI to learn and adapt in real-time to unforeseen environmental changes, potentially integrating unsupervised and meta-learning techniques, broadening autonomous drone applications in challenging conditions.

\addtolength{\textheight}{-8cm} 






\begin{thebibliography}{99}

\bibitem{b0} M.Bajpai, P.Singh and S.Sinha, “A review on the applications of UAVs in various field , “JETIR, Volume 6, Issue 5 ,2019.


\bibitem{b1}S. Lee, T. Shim, S. Kim, J. Park, K. Hong and H. Bang, “Vision-Based Autonomous Landing of a Multi-Copter Unmanned Aerial Vehicle using Reinforcement Learning," International Conference on Unmanned Aircraft Systems (ICUAS), 2018, pp. 108-114.

\bibitem{b2} R. Polvara, M.Patacchiola, S.Sharma1, J.Wan1, A.Manning, R.Sutton and A. Cangelosi “Toward End-to-End Control for UAV Autonomous Landing via Deep Reinforcement Learning," International Conference on Unmanned Aircraft Systems (ICUAS), 2018, pp. 115-123.

\bibitem{b3} M. B. Vankadari, K. Das, C. Shinde and S. Kumar, “A Reinforcement Learning Approach for Autonomous Control and Landing of a Quadrotor," International Conference on Unmanned Aircraft Systems (ICUAS), 2018, pp. 676-683.

\bibitem{b4} S. Karaf, A. Fedoseev, M. Martynov, Z. Darush, A. Shcherbak and D. Tsetserukou, “MorphoLander: Reinforcement Learning Based Landing of a Group of Drones on the Adaptive Morphogenetic UAV," IEEE International Conference on Systems, Man, and Cybernetics (SMC), 2023, pp. 2507-2512.

\bibitem{b5} J. Amendola, L. R. Cenkeramaddi and A. Jha, “Single Reinforcement Learning Policy for Landing a Drone Under Different UGV Velocities and Trajectories," International Conference on Control, Mechatronics and Automation (ICCMA), 2023, pp. 115-120.

\bibitem{b6} Z. Jiang and G. Song, “A Deep Reinforcement Learning Strategy for UAV Autonomous Landing on a Platform," International Conference on Computing, Robotics and System Sciences (ICRSS),2022, pp. 104-109.

\bibitem{b7} J.Tsai, P.Chen,M.Tsai, “Accuracy Improvement of Autonomous Straight Take off,Flying Forward, and Landing of a Drone with Deep Reinforcement Learning," ICCETW,2019.

\bibitem{b8} Y. Rao, S. Ma, J. Xing, H. Zhang and X. Ma, “Real time vision-based autonomous precision landing system for UAV airborne processor," Chinese Automation Congress (CAC), 2020, pp. 532-537.

\bibitem{b9} T. Swain, M. Rath, J. Mishra, S. Banerjee and T. Samant, “Deep Reinforcement Learning based Target Detection for Unmanned Aerial Vehicle,"India Council International Subsections Conference (INDISCON), Bhubaneswar, 2022, pp. 1-5.

\bibitem{b10} J. Bialas and M. Doller, “Coverage Path Planning for Unmanned Aerial Vehicles in Complex 3D Environments with Deep Reinforcement Learning," IEEE International Conference on Robotics and Biomimetics (ROBIO), 2022, pp. 1080-1085.

\bibitem{b11} M. Piponidis, P. Aristodemou and T. Theocharides, “Towards a Fully Autonomous UAV Controller for Moving Platform Detection and Landing," International Conference on VLSI Design and 2022 21st International Conference on Embedded Systems (VLSID), 2022, pp. 180-185.

\bibitem{b12} F.Hauf, B.Kocer, A.Slatter, H.Nguyen, O.Pang, R.Clark, E.Johns and M.Kovac1, “Learning Tethered Perching for Aerial Robots," IEEE International Conference on Robotics and Automation (ICRA), 2023, pp. 1298-1304.

\bibitem{b13} K. Backman, D. Kulic and H. Chung, “Learning to Assist Drone Landings," IEEE Robotics and Automation Letters, vol. 6, no. 2, pp. 3192-3199, 2021.

\bibitem{b14} Y. Song, M. Steinweg, E. Kaufmann and D. Scaramuzza, “Autonomous Drone Racing with Deep Reinforcement Learning," IEEE/RSJ International Conference on Intelligent Robots and Systems (IROS), 2021, pp. 1205-1212.

\bibitem{b15} J. Panerati, H. Zheng, S. Zhou, J. Xu, A. Prorok and A. P. Schoellig, “Learning to Fly—a Gym Environment with PyBullet Physics for Reinforcement Learning of Multi-agent Quadcopter Control," IEEE/RSJ International Conference on Intelligent Robots and Systems (IROS), 2021, pp. 7512-7519.

\bibitem{b16} G. Shi, X.Shi,M.Connell, R.Yu,K. Azizzadenesheli, A.Anandkumar, Y.Yue, S.Chung, “Neural Lander: Stable Drone Landing Control Using Learned Dynamics," 2019 International Conference on Robotics and Automation (ICRA), 2019, pp. 9784-9790.

\bibitem{b17} Z.Feiyu, L.Dayan, W.Zhengxu, M.Jianlin and W.Niya, “Autonomous localized path planning algorithm for UAVs based on TD3 strategy," Sci Rep 14, 763, 2024.

\bibitem{b18} K.Backman, D.Kulic and H.Chung, “ Reinforcement learning for shared autonomy drone landings," Auton Robot 47,2023, pp. 1419-1438.

\bibitem{b21} A.Keipour, A.S. Pereira, R.Bonatti, R.Garg, P.Rastogi, G.Dubey and S. Scherer, “Visual Servoing Approach to Autonomous UAV Landing on a Moving Vehicle ,”Sensors, 22(17), 6549,2022.

\bibitem{b22} E.Kaufmann, L.Bauersfeld, A. Loquercio, M.Muller, V.Koltun and D.Scaramuzza, “Champion-level drone racing using deep reinforcement learning," Nature,620, 2023, pp, 982–987.

\bibitem{b23} G.Ting,J.Gau, “UAV Path Planning and Obstacle Avoidance Based on Reinforcement Learning in 3D Environments,”Special Issue Intelligent Control and Robotic System in Path Planning,MDPI,2023.

\bibitem{b24} G. Xu, L. Liu and H. Liu, “UAV Landing Control Disturbed by Carrier Air-Wake Based on Deep Reinforcement Learning," 34th Chinese Control and Decision Conference (CCDC), 2022, pp. 4719-4724.

\bibitem{b25} M.Tahir, I.Mir and T.Islam, “Control Algorithms, Kalman Estimation and Near Actual Simulation for UAVs: State of Art Perspective," Drones,7(6),339,MDPI,2023.

\bibitem{b26} N.Lin, D.Zhao, M.Sellier and X.Liu, “Experimental investigation on turbulence effects on unsteady aerodynamics performances of two horizontally placed small-size UAV rotors,"Aerospace Science and Technology,141,2023.

\bibitem{b27} J. E. Kooi and R. Babuska, “Inclined Quadrotor Landing using Deep Reinforcement Learning," IEEE/RSJ International Conference on Intelligent Robots and Systems (IROS), 2021, pp. 2361-2368.

\bibitem{b27}
A. Raffin, A. Hill, A. Gleave, A. Kanervisto, M. Ernestus, and N. Dormann,
“Stable-Baselines3: Reliable Reinforcement Learning Implementations,"
\emph{Journal of Machine Learning Research}, vol. 22, no. 268, pp. 1-8, 2021.
\bibitem{b28}
A. Gupta, E. Dorzhieva, A. Baza, M. Alper, A. Fedoseev and D. Tsetserukou, "SwarmHawk: Self-Sustaining Multi-Agent System for Landing on a Moving Platform through an Agent Supervision," 2022 International Conference on Unmanned Aircraft Systems (ICUAS), Dubrovnik, Croatia, 2022, pp. 990-997, doi: 10.1109/ICUAS54217.2022.9836080.
\end{thebibliography}
\end{document}